%% file: ms.tex
\documentclass[letterpaper, 10 pt, conference]{ieeeconf}  

\IEEEoverridecommandlockouts                              

\title{\LARGE \bf
Robust Model-free Reinforcement Learning with Multi-objective Bayesian Optimization
}

\author{Matteo Turchetta$^{1}$ Andreas Krause$^{1}$ Sebastian Trimpe$^{2}$
\thanks{$^{1}$ Learning and Adaptive Systems Group, ETH Zurich, Z\"{u}rich, Switzerland
        {\tt\small matteotu@inf.ethz.ch krausea@ethz.ch}}%
\thanks{$^{2}$ Intelligent Control Systems Group,
        Max Planck Institute for Intelligent Systems, Stuttgart, Germany
        {\tt\small trimpe@is.mpg.de}}
\thanks{This research was supported in part by the Max Planck ETH Center for
Learning  Systems,  the Max Planck Society and the Cyber Valley Initiative.}
}

\usepackage{amsmath}
\usepackage{amsfonts}
\usepackage[capitalise]{cleveref}
\usepackage{bm}
\usepackage{todonotes}

\usepackage{layouts}

\usepackage{algorithm}
\usepackage[noend]{algorithmic}

\usepackage{booktabs}
\usepackage{multirow}
\usepackage{siunitx}

\input{defs.tex}

\begin{document}

\maketitle
\thispagestyle{empty}
\pagestyle{empty}

\input{sections/0-abstract.tex}
\input{sections/1-introduction.tex}
\input{sections/2-problem_statement.tex}
\input{sections/3-method.tex}
\input{sections/5-experiments.tex}
\input{sections/6-conclusions.tex}

\bibliographystyle{IEEEtran}
\bibliography{IEEEabrv,icra20}



\end{document}

%% file: defs.tex
\newcommand{\perf}{f_1}
\newcommand{\rob}{f_2}
\newcommand{\sys}{h}

\newcommand{\dom}{\Theta}
\newcommand{\param}{\theta}

\newcommand{\x}{\mathbf{x}}
\newcommand{\y}{\mathbf{y}}
\newcommand{\obs}{\mathbf{o}}
\newcommand{\f}{\mathbf{f}}
\newcommand{\ub}{\mathbf{u}}
\newcommand{\vb}{\mathbf{v}}
\newcommand{\wb}{\mathbf{w}}

\newcommand{\simtoreal}{sim-to-real }

\crefname{equation}{}{} 
\crefname{section}{Sec.}{Sec.}
\crefname{figure}{Fig.}{Fig.}
\crefname{lemma}{Lemma}{Lemma}

\DeclareMathOperator*{\argmaxaux}{argmax}
\newcommand{\argmax}{\displaystyle\argmaxaux}

%% file: sections/0-abstract.tex
\begin{abstract}
    In reinforcement learning (RL), an autonomous agent learns to perform complex tasks by maximizing an exogenous reward signal while interacting with its environment. In real world applications, test conditions may differ substantially from the training scenario and, therefore, focusing on pure reward maximization during training may lead to poor results at test time. In these cases, it is important to trade-off between performance and robustness while learning a policy. While several results exist for robust, model-based RL, the model-free case has not been widely investigated. In this paper, we cast the robust, model-free RL problem as a multi-objective optimization problem. To quantify the robustness of a policy, we use delay margin and gain margin, two robustness indicators that are common in control theory. We show how these metrics can be estimated from data in the model-free setting. We use multi-objective Bayesian optimization (MOBO) to solve efficiently this expensive-to-evaluate, multi-objective optimization problem. We show the benefits of our robust formulation both in \simtoreal and pure hardware experiments to balance a Furuta pendulum.
\end{abstract}

%% file: sections/1-introduction.tex
\section{Introduction} \label{sec:introduction}
In reinforcement learning (RL) \cite{sutton2018reinforcement}, the goal is to learn a controller to perform a desired task from the data produced by the interaction between the learning agent and its environment. In this framework, autonomous agents are trained to maximize their return. It is common to assume that such agents will be deployed in conditions that are similar, if not equal, to those they were trained in. In this case, a return-maximizing agent performs well at test time. However, in real world applications, this assumption may be violated. For example, in robotics, we can use RL to learn to fly a drone indoor. However, later on we may use the same drone to carry a payload in a windy environment. The new environmental conditions and the possible deterioration of the drone components due to their usage may result in a poor, if not catastrophic, performance of the learned controller. Another scenario where training and testing conditions differ substantially is the \simtoreal setting, i.e., when we deploy a controller trained in simulation on a real-world agent.

Considering robustness alongside performance when learning a controller can limit performance degradation due to different training and testing environments. In special cases, these goals may be aligned, and a high-performing controller can also be robust. This is the case for the Linear Quadratic Regulator (LQR), a linear state-feedback controller that is optimal for the case of linear dynamics, quadratic cost, and perfect state measurements.  It is well-known that the LQR exhibits strong robustness indicators, such as gain and phase margins \cite{anderson2007optimal}.  
While performance and robustness go hand in hand for the LQR, they are often conflicting in other cases. For example, a celebrated result in control theory shows that the Linear Quadratic Gaussian (LQG) regulator - the noisy counterpart of the LQR - 
can be arbitrarily close to instability, despite being optimal \cite{doyle1978guaranteed}. Thus, in general, we need to trade-off between performance and robustness \cite{boulet2007fundamental}.

\textbf{Contributions.} While many works investigating the performance/robustness trade-off exist in both the RL and control theory literature for the model-based setting, few results are known for the model-free scenario. However, there are several real-world scenarios where models are not available, inaccurate, or too expensive to use, but robustness is fundamental. Thus, in this paper, we introduce the first data-efficient, robust, model-free RL method based on policy optimization with multi-objective Bayesian optimization (MOBO). In particular, these are our contributions:
\begin{itemize}
    \item We formulate the robust, model-free RL as a multi-objective optimization problem.
    \item We propose a model-free, data-driven evaluation of delay and gain margins, two common robustness indicators from the model-based setting (where they are computed analytically).
    \item We solve this problem efficiently with expected hypervolume improvement (EHI). 
    \item We introduce the first method that can learn robust controllers directly on hardware in a model-free fashion.
    \item We show how our approach outperforms non-robust policy optimization in evaluations on a Furuta pendulum for both a \simtoreal and a pure hardware setting.
\end{itemize}

\textbf{Related work.} Robustness has been widely investigated in control theory \cite{zhou1998essentials}, and standard robust control techniques for linear systems include loop transfer recovery \cite{saberi1993loop}, $H_\infty$ control, and $\mu$ synthesis \cite{zhou1998essentials,green2012linear}. However, these methods typically assume the availability of a model, and none of these includes a learning component. Recently, robustness has drawn attention in data-driven settings, giving rise to the field of robust, model-based RL. 
Robust Markov decision processes study the RL problem when the transition model is subject to known and bounded uncertainties. For example, \cite{nilim2005robust} studies the dynamic programming recursion in this setting. Other methods that consider parametric uncertainties include \cite{sharma2007robust,delage2010percentile}. All the previous methods are model-based. 

Robustness and performance are typical objectives in control design, which often conflict each other, thus requiring design trade-offs \cite{boulet2007fundamental,astrom2008feedback}. In the model free literature, this trade-off is often fixed a priori and the resulting problem is solved with standard optimization methods. In \cite{NeuMarSchTri18} a weighted cost that balances performance and robustness is optimized. In \cite{venkataraman2019recovering} robust controllers are learned via gradient ascent with random multiplicative noise on the control action. In \cite{pinto2017robust,morimoto2001robust} external, adversarial disturbances are used instead. In these works, the upper bound on the magnitude of the disturbance implicitly balances robustness and performance. However, setting this trade-off is often not intuitive and, in case the requirements are misspecified or updated, a new controller must be learned. Alternatively, robust control design methods based on multi-objective optimization explore the spectrum of such trade-offs. The work in \cite{gambier2011multi} gives a review of such methods, with a focus on genetic algorithms, which, due to their low data efficiency require the model to compute the robustness indices.

Model-free RL algorithms are typically validated in simulations due to their high sample complexity. However, in robotics, it is crucial to test these methods on hardware. Bayesian optimization (BO) \cite{mockus1978application,shahriari2016taking} has been successfully applied to learn low-dimensional controllers for hardware systems. 
For example, \cite{berkenkamp2016safe} learns to control the $x$-coordinate for a quadrotor hovering task with a linear controller, \cite{marco2017virtual} learns a linear state feedback controller for a cart-pole system in a \simtoreal setting and \cite{Calandra2016,antonova2017deep} tune the parameters of ad-hoc controllers for locomotion tasks. However, none of these methods considers robustness, making ours the first one to learn robust controllers from data directly on hardware.

MOBO is the branch of BO that solves multi-objective problems. MOBO algorithms include EHI \cite{Emmerich2008}, $\epsilon$-PAL \cite{Zuluaga2016}, and PESMO  \cite{hernandez2016predictive}. They have been applied to several tasks including trading off prediction speed and accuracy in machine learning models.
However, they have rarely been applied to RL. To the best of our knowledge, this has been done only in \cite{Tesch2013ExpensiveMO, Ariizumi2014}, where a trade-off between frontal camera movement and forward speed is found for a snake-like robot, for homoschedastic and heteroschedastic noise respectively.  Robustness is not explicitly treated in these works.







\begin{figure}[t]
    \centering
    \includegraphics[scale=.33]{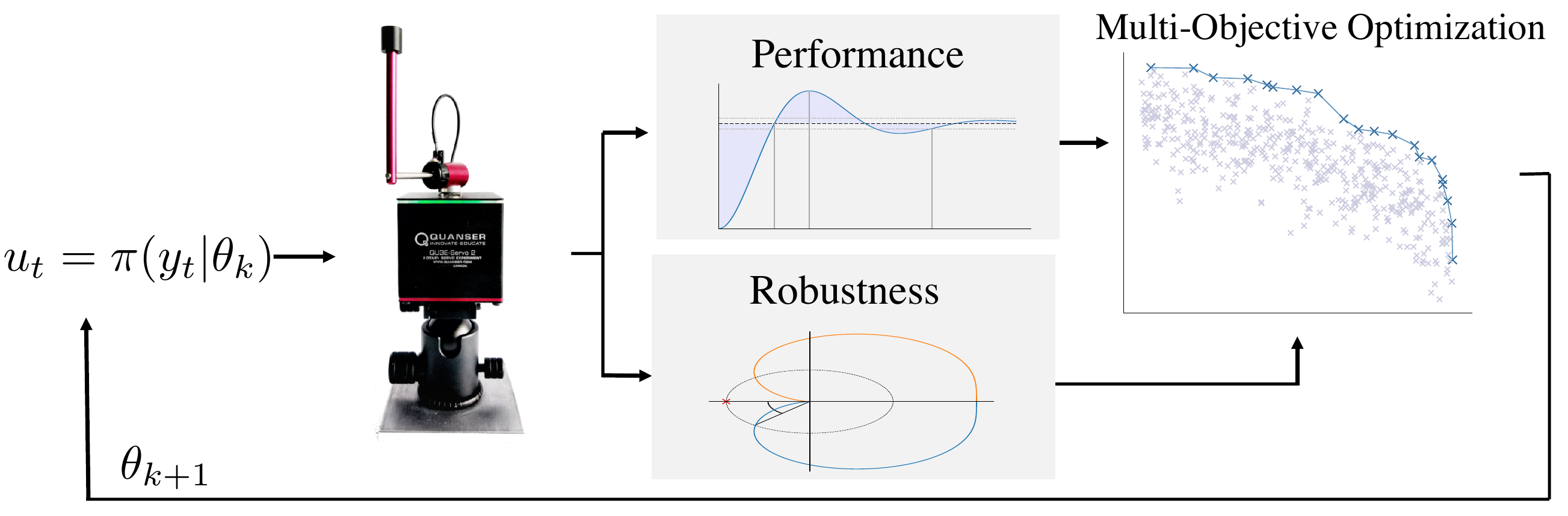}
    \caption{The multi-objective optimization model for the robust, model-free RL problem. We choose a controller corresponding to parameters $\theta_k$. Subsequently, we evaluate its performance and robustness by deploying it on the system. Finally, we use these observations to choose a new controller.}
    \label{fig:MOBO_loop}
    \vspace{-.35cm}
\end{figure}


%% file: sections/2-problem_statement.tex
\section{Problem Statement}
In this section, we introduce our formulation of robust model-free RL as a multi-objective optimization problem. For ease of exposition, we limit ourselves to two objectives. However, this approach naturally extends to the any number of objectives, for example, multiple robustness indicators.

We assume we have a system with \emph{unknown} dynamics, $h$, and \emph{unknown} observation model, $g$, 
\begin{equation}
    \x_{t+1} = \sys (\x_t,\ub_t,\wb_t),  ~~~~~
    \obs_t = g(\x_t,\vb_t) \label{eq:sys}
\end{equation}
where $\x$ is the state, $\ub$ is the control input, $\obs$ is the observation and $\wb$ and $\vb$ are the process and sensor noise. An RL agent aims at learning a controller $\ub_t=\pi(\obs_t|\param)$, i.e., a mapping parametrized by $\param$ from an observation $\obs_t$ to an action $\ub_t$ that allows it to complete its task. Policy optimization algorithms are a class of model-free RL methods that solve this problem by optimizing the performance of a given controller for the task at hand as a function of the parameters $\param$. Concretely, given a performance metric $f_1:\dom\rightarrow \mathbb{R}$, standard, non-robust policy optimization algorithms aim to find $\param^* \in \argmax~f_1(\param)$.
In this work, we consider regulation tasks, i.e., bringing and keeping the system in a desired goal state $\overline{\x}$. This includes common problems like stabilization, set-point tracking, or disturbance rejection. The performance indicator $f_1$ encodes these objectives.

To extend this framework to the robustness-aware case, we use a second function $f_2:\dom\rightarrow \mathbb{R}$ that measures the robustness of a controller. Since both the dynamics $h$ and the observation model $g$ are unknown, we must evaluate or approximate the value of $f_2$ from data. In \cref{sec:robustness}, we introduce the gain and the delay margin, two alternatives for $f_2$ that are commonly used in model-based control and we discuss how to evaluate them in the model-free setting.

We aim at finding the best controller in terms of  performance and robustness, as measured by $\perf$ and $\rob$. However, since we compare controllers based on multiple, and possibly conflicting, criteria, we cannot define a single best controller. Given a controller $\param$, we denote with $\f_\param=[\perf(\param), \rob(\param)]$ the array containing its performance and robustness values.  To compare two controllers $\param_1$ and $\param_2$, we use the canonical partial order over $\mathbb{R}^2$: $\f_{\param_1}\succeq \f_{\param_2}$ iff $f_i(\param_1)\geq f_i(\param_2)$ for $i=1, 2$. This induces a relation in the controller space $\dom$: $\param_1 \succeq \param_2$ iff $\f_{\param_1}\succeq \f_{\param_2}$. If $\param_1 \succeq \param_2$, we say that $\param_1$ dominates $\param_2$. The Pareto set $\dom^*\subseteq \dom$ is the set of non-dominated points in the domain, i.e., $\param^* \in \dom^*$ iff $\exists i=1, 2$ such that $f_i(\param^*)>f_i(\param)$ for all $\param \in \dom$. The Pareto front is the set of function values corresponding to the Pareto set. The Pareto set is optimal in the sense that, for each  point $\param^*$ in it, it is not possible to find another point in the domain that improves the value of one objective without degrading another \cite{collette2013multiobjective}. The goal of this paper is to approximate $\dom^*$ from data. 

\cref{fig:MOBO_loop} represent our problem graphically: we suggest a controller, we evaluate its performance and robustness on the system and we select a new controller based on these observations to find an approximation of the Pareto front.

%% file: sections/3-method.tex
\section{Learning the Performance-robustness Trade-off} \label{sec:method}
For the robust, model-free RL setting we consider, we propose to learn the Pareto front characterizing the performance-robustness trade-off of a given system with MOBO.
Here, we describe the necessary components to solve our problem in a data efficient way: MOBO and the robustness and performance indicators used in our experiments. Moreover, we discuss how to evaluate such indicators from data in a model-free fashion.
\subsection{Multi-objective Bayesian optimization}
MOBO algorithms solve multi-objective optimization problems by sequentially querying the objective at different inputs and obtaining noisy evaluations of the corresponding values. They build a statistical model of the objectives to capture the belief over them given the data available. They measure how informative a point in the domain is about the problem solution with an acquisition function. At every iteration, they evaluate the objective at the most informative point, as measured by the acquisition function. Thus, the complex multi-objective optimization problem is decomposed into a sequence of simpler scalar-valued optimization problems. In the following, we describe the surrogate model and the acquisition function used in this work.

\textbf{Intrinsic Model of Coregionalization}
A single-output Gaussian process (GP) \cite{rasmussen2004gaussian} is a probability distribution over the space of functions of the form $f:\dom \rightarrow \mathbb{R}$, such that the joint distribution of the function values computed over any finite subset of the domain follows a multi-variate Gaussian distribution. A GP is fully specified by a mean function $\mu:\dom \rightarrow \mathbb{R}$, which, w.l.o.g., is usually assumed to be zero, $\mu(\param)=0$ for all $\param \in \dom$, and a covariance function, or kernel, $k:\dom\times\dom \rightarrow \mathbb{R}$. The kernel encodes the strength of statistical correlation between two latent function values and, therefore, it expresses our prior belief about the function behavior.

Similarly, a $D$-output GP is a probability distribution over the space of functions of the form $\f:\dom \rightarrow \mathbb{R}^D$. The difference with respect to  single-output GPs is that, in this case, the kernel must capture the correlation across different output dimensions in addition to the correlation of function values at different inputs. The simplest way of doing this is by assuming that each output is independent. However, this model disregards the fundamental trade-off between robustness and performance that we are considering. For a review on kernels for multi-output GPs, see \cite{alvarez2012kernels}. In this work, we use the intrinsic model of coregionalization (ICM), which defines the covariance between the $i^\text{th}$ value of $\f(\param)$ and the $j^\text{th}$ value of $\f(\param^\prime)$ by separating the input and the output contribution as follows, $b_{ij}k(\param, \param^\prime)$. In this case, we say $\f\sim\mathcal{GP}(\boldsymbol{\mu}(\cdot), \mathbf{K}(\cdot, \cdot)=\mathbf{B}k(\cdot, \cdot))$, where $\boldsymbol{\mu}:\dom\rightarrow \mathbb{R}^D$ is a $D$-dimensional mean function, $k:\dom\times\dom \rightarrow \mathbb{R}$ is a scalar-valued kernel and $\mathbf{B}\in \mathbb{R}^{D\times D}$ is a matrix describing the correlation in the output space (more details on $\mathbf{B}$ in \cref{sec:experiments}).
Given $N$ noisy observations of $\f$, $\mathcal{D}=\{(\param_1, \y_1), \cdots\,(\param_N, \y_N) \}$, with $\y_i = \f(\param_i)+\omega_i$, where $\omega_i \sim \mathcal{N}(\mathbf{0}, \Sigma)$ is i.i.d. Gaussian noise, we can compute the posterior distribution of the function values conditioned on $\mathcal{D}$ at a target input $\param^\star$ in closed form as $p(\f(\param^\star)|\mathcal{D},\param^\star)\sim\mathcal{N}(\f^\star(\param^\star), \mathbf{K}^\star(\param^\star,\param^\star)))$. We denote with $\boldsymbol{\param}$ the inputs contained in $\mathcal{D}$ and with $\mathbf{K}(\boldsymbol{\param},\boldsymbol{\param})$ the $ND\times ND$ matrix with entries $(\mathbf{K}(\param_u, \param_v))_{i,j}$ for $u,v=1, \ldots,N$ and $i,j=1,\ldots,D$, then
\begin{align}
    &\f^\star(\param^\star)=\mathbf{K}_{\param^\star}^\top(\mathbf{K}(\boldsymbol{\param},\boldsymbol{\param})+\boldsymbol{\Sigma})^{-1}\overline{\y},\\
    &\mathbf{K}^\star(\param^\star, \param^\star) = \mathbf{K}(\param^\star, \param^\star) - \mathbf{K}_{\param^\star}(\mathbf{K}(\boldsymbol{\param},\boldsymbol{\param})+\boldsymbol{\Sigma})^{-1}\mathbf{K}_{\param^\star}^\top,
\end{align}
where $\boldsymbol{\Sigma}=\Sigma\otimes I_N$, with $\otimes$ denoting the Kronecker product, $\mathbf{K}_{\param^\star}\in \mathbb{R}^{D\times ND}$ has entries $(\mathbf{K}(\param^\star, \param_v))_{i,j}$ for $v=1, \ldots,N$ and $i,j=1,\ldots,D$ and $\overline{\y}$ is the $ND$-dimensional vector containing the concatenation of the observations in $\mathcal{D}$.


\textbf{Expected Hypervolume Improvement}
EHI is an acquisition function introduced in \cite{Emmerich2008}, which selects inputs to evaluate based on a notion of improvement with respect to the incumbent solution. In multi-objective optimization, incumbent solutions take the form of approximations of the Pareto set, $\mathcal{X}^*$, whose quality is measured by the hypervolume indicator induced by the corresponding front, $\mathcal{Y}^*$ with respect to a reference $r$. Formally, the hypervolume indicator of a set of points $A$ with respect to a reference $r$, $\text{HV}(A;r)$, is the Lebesgue measure of the hypervolume covered by the boxes that have an element in $A$ as upper corner and the reference as lower corner. It quantifies the size of the portion of the output space that is Pareto-dominated by the points in $A$. Given an estimate of the Pareto front, $\mathcal{Y}^*$, the hypervolume improvement of $\param \in \dom$ is defined as the relative improvement in hypervolume obtained by adding the function value at $\param$, $f(\param)$, to $\mathcal{Y}^*$, $\text{HI}(f(\param);\mathcal{Y}^*,r)=\text{HV}(\mathcal{Y}^*\cup \f(\param);r) - \text{HV}(\mathcal{Y}^*;r).$   
However, we do not know $\mathbf{f}(\param)$. Instead, we have a belief over its value expressed by the posterior distribution of the GP, which, in turn, induces a distribution over the hypervolume improvement corresponding to an input $\param$. The EHI acquisition function quantifies the informativeness of an input $\param$ toward the solution of the multi-objective optimization problem through the expectation of this distribution,
\begin{equation}
    \alpha(\param|\mathcal{D}, \mathcal{Y}^*, r)=\int_{f(\param) \in \mathbb{R}^n}\text{HI}(\f(\param);\mathcal{Y}^*), r)p(\f(\param)|\mathcal{D})d\f(\param). \label{eq:EHI}
\end{equation}
\cite{Emmerich2008} shows how to compute the integral in \cref{eq:EHI} in closed form.


\subsection{Robustness}\label{sec:robustness}
œ¬In general, robustness can have very different meanings. One may desire to ensure robustness to a certain class of disturbances, imperfections in the control system, or uncertainty in the process, for example. In control theory, the latter is often understood as robustness in the stricter sense.  Specifically, \emph{robust stability} assures that a controller stabilizes every member from a set of uncertain processes \cite{zhou1998essentials}.  Such processes can, for example, be defined through a nominal process and variations thereof. Different variations lead to different robustness characterizations.  Likewise, there are different notions of stability that are meaningful depending on the context. For example, for a deterministic system, asymptotic stability, i.e., $\x_t \to \overline{\x}$ as $t \to \infty$, where $\overline{\x}$ is an equilibrium of the system, is often used; for systems that are continuously excited, e.g., through noise, and thus cannot approach $\overline{\x}$, one may seek 
the above limit to hold in expectation or practical stability in the sense of a bounded state, i.e., $\|\x_t-\overline{\x}\| \leq \x_\mathrm{max}$ for all $t\geq \overline{t}$. A controller is unstable when the respective condition does not hold (e.g., no asymptotic convergence, or $\x_t$ grows beyond any bounds).  

While many sophisticated robustness metrics have been developed, stability margins such as gain and delay margins are some of the most  common and intuitive ones \cite[Sec.~9.3]{astrom2008feedback}. We consider these in this work and comment on alternatives in \cref{sec:conclusion}.
Below, we formally introduce them and we explain how to evaluate them in a model-free setting. Notice that, our data-driven definitions can be  extended to any setting where a success/failure outcome can be defined and, therefore, are not limited to stability considerations.

\textbf{Gain margin.} 
In classical control, the \emph{upper} (\emph{lower}) \emph{gain margin} is defined for single-input-single-output (SISO) linear systems as the largest factor $\kappa_\text{max} \in (1, \infty)$ (the smallest factor $\kappa_\text{min} \in (0,1)$) that can multiply the open-loop transfer function so that the closed-loop system is stable \cite[Sec.~9.5]{zhou1996robust}.  As the open-loop transfer function encodes both the process and the controller dynamics, the factor may represent uncertainty in the process gain or the actuator efficiency, for example. 
In this work, we consider a factor $\kappa$ to be multiplied by the control action (i.e., $\ub_t = \kappa \times \pi(\obs_t|\param)$), which is equivalent to the definition for linear SISO systems, but can also be used for nonlinear ones.
It quantifies how much we can lower/amplify the control action before making the system unstable. 
In a way, it quantifies how ``far'' we are from instability and, thus, how much we can tolerate differences between training and testing.


\textbf{Delay margin.}
Similarly, we define the \emph{delay margin} as the largest time delay on the measurement $\obs_t$ such that the controlled system is still stable. 
Formally, it is the largest value of $d\in(0, \infty)$ such that the closed-loop system with the delayed control action $\ub_t = \pi(\obs_{t-d}|\theta)$ is stable.  As delay in data transmission between sensor, controller, and actuator, and in the control computation are present in most control systems, the delay margin is a very relevant measure. 


\textbf{Estimate from data.}
While the indicators above can be readily computed for linear systems, they are difficult to compute analytically if the model is nonlinear, or impossible if no model is available, as considered herein.
We describe an experiment to estimate the delay margin from data in a model-free setting (those for the gain margins are analogous). For general non-linear systems, stability with respect to an equilibrium is a local property. Thus, we assume we can reset the system to a state in the neighborhood of the equilibrium of interest, i.e., we have $\x_0 \in \mathcal{B}(\overline{\x}, \rho)$, where $\mathcal{B}(\overline{\x}, \rho)$ is a ball centered at $\overline{\x}$ of radius $\rho$. 
We can establish whether the delay margin, denoted with $d^*$, is larger or smaller than a delay $d$ by resetting the system near $\overline{\x}$, deploying the delayed controller $\pi(\obs_{t-d}|\param)$ and evaluating the stability of the resulting trajectory. 

\begin{algorithm}[t]
    \caption{Robust policy optimization}
    \label{alg:EHI}
    \begin{algorithmic} [1]
        \STATE \textbf{Inputs}: Reference $r$
        \STATE $\mathcal{D}_0\gets\emptyset,~\mathcal{Y}_0^*\gets\{r\}$
        \FOR{$k=0, 1, \ldots$}
        \STATE $\param_{k+1} \gets \textrm{argmax}_{\param \in \dom}~\textrm{EHI}(\param|\mathcal{D}_k, \mathcal{Y}^*_k, r)$ 
        \STATE $y_{k+1}^1 \gets \text{\tt performance}\_\text{\tt experiment}(\param_{k+1})$
        \STATE $y_{k+1}^2 \gets \text{\tt robustness}\_\text{\tt experiment}(\param_{k+1})$
        \STATE $\mathcal{D}_{k+1} \gets \mathcal{D}_k \cup \{(\param_{k+1}, [y_{k+1}^1, y_{k+1}^2])\}$
        \STATE $\dom^*_{k+1}, \mathcal{Y}^*_{k+1} \gets \text{\tt Pareto}\_\text{\tt set}\_\text{\tt and}\_\text{\tt front}(\mathcal{D}_{k+1})$
    \ENDFOR
    \STATE \textbf{Outputs}: Pareto set $\dom^*$, Pareto front $\mathcal{Y}^*$
    \end{algorithmic}
\end{algorithm}

\begin{figure*}
    \centering
    \includegraphics[width=.85\textwidth]{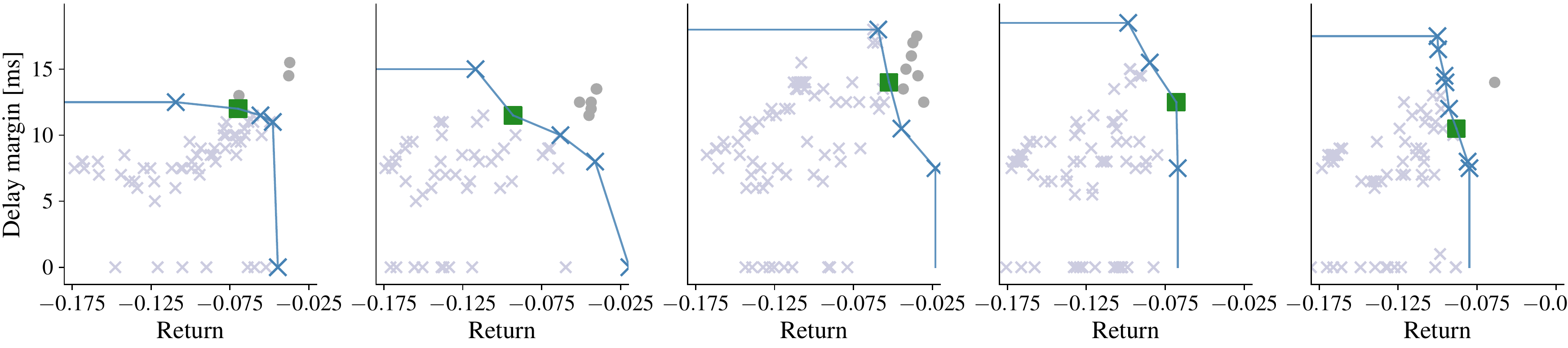}
    \caption{Pareto fronts identified in simulation for the delay margin experiment (MOBO-DM). The green square indicates the controller tested on the hardware, see \cref{tab:sim2real}. The gray circles indicate the controllers that appeared to be outliers and were discarded after running longer simulations.}
    \label{fig:dm_front}
\end{figure*}

\begin{figure*}
    \centering
    \includegraphics[width=.85\textwidth]{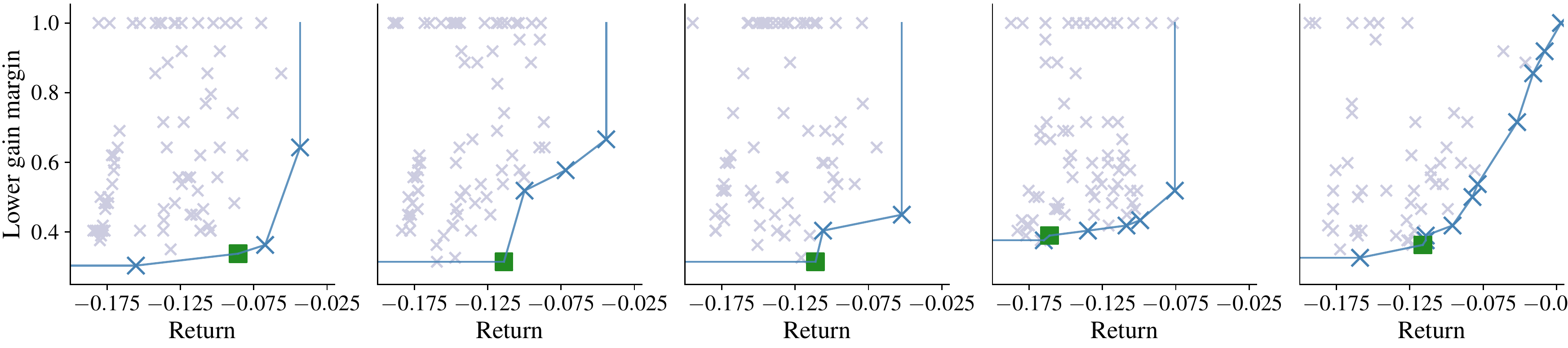}
    \caption{Pareto fronts identified in simulation for the gain margin experiment (MOBO-GM). The green square indicates the controller that is tested on the hardware, see \cref{tab:sim2real}.}
    \label{fig:gm_front}
\end{figure*}

In practice, two problems arise  with this approach: \emph{(i)} we can evaluate a finite number of delays with a finite number of experiments; and \emph{(ii)} while stability is an asymptotic condition on the state, we do not know the state and we run finite experiments. The first problem requires us to select carefully the delays we evaluate. We know that increasing values of delay take a stable system closer to instability. Thus, given $m$ delays $[d_1, \cdots, d_m]$, we do a binary search to find the largest one for which the closed-loop system is stable. This allows us to approximate the delay margin with $\log m$ experiments. 
Not knowing the state $\x$ can be solved by estimating it from the noisy sensor measurements $\obs_t$ or by introducing a new definition of stability based on $\obs_t$ rather than $\x_t$. Concerning the finite trajectories we note that, in practical cases, it is rare to have small compounding deviations from $\overline{\x}$  resulting in a divergent behavior emerging only in the long run. Often, a controller makes the system converge to or diverge from $\overline{\x}$ within a short amount of time.
In our experiments, we say that a controller stabilizes the system if, after a burn-in time that accounts for the transient behavior, it keeps the state within a box around $\overline{\x}$. Controllers with good margins are investigated further with longer experiments to eliminate potential outliers due to the finite trajectory issue.

Reliably estimating robustness indicators and stability of a system without a model is challenging. The estimation technique we presented is intuitive and easy to implement. While it does not provide formal guarantees on the estimation error, we show in \cref{sec:experiments} that it is accurate enough to greatly improve the robustness of our algorithm with respect to the non-robust policy optimization baseline.


\subsection{Algorithm}  \label{sec:algorithm}
In this section, we describe our robust policy optimization algorithm, for the pseudocode see \cref{alg:EHI}. At iteration $k$, we select the controller that maximizes the EHI criterion. Then, we run two experiments to estimate its performance and robustness. For the performance, we introduce a state and action dependent reward and we define the return as the average reward obtained over an episode. The performance index is defined as the expectation of the return, which we approximate with a Monte Carlo estimate over multiple episodes. To estimate the robustness, we use the experiments from \cref{sec:robustness}. We update the data set with the experiments results. Finally, we update the estimate of the Pareto front that is used to compute the EHI as the set of dominating points of the data set. Other options to compute such estimate from the posterior of the GP exist. However, they are computationally more expensive and they resulted in a similar performance in our experiments. In the end, the algorithm returns an estimate of the Pareto set and front. The choice of a controller from the Pareto set depends on the performance-robustness trade-off required by the test applications and, therefore, the choice is left to the practitioner.

%% file: sections/5-experiments.tex
\section{Experimental Results} \label{sec:experiments}
%

We compare the robust policy optimization algorithm in \cref{alg:EHI} to its non-robust counterpart based on scalar BO as, e.g., in \cite{berkenkamp2016safe,marco_ICRA_2016}. We use the scalar equivalent of EHI for the non-robust case, i.e., the expected improvement (EI) algorithm \cite{mockus1978application}. We present two set of experiments: training controllers in simulation and directly on hardware, respectively. In both cases, the learned controllers are tested on the hardware in a set of different conditions.

\textbf{System.} We learn a controller for a Furuta pendulum \cite{cazzolato2011dynamics} (see Figure~\ref{fig:MOBO_loop}), a system that is closely related to the well-known cart pole. It replaces the cart with a rotary arm that  rotates on the horizontal plane. In our experiments, we use the Qube Servo 2 by Quanser \cite{QubeManual},  a small scale Furuta pendulum. It uses a brushless DC motor to exert a torque on the rotary arm, and it is equipped with two incremental optical encoders with 2048 counts per revolution  to measure the angle of the rotary arm and the pendulum. For sim-to-real, we use the dynamics model provided in the Qube Servo 2 manual \cite{QubeManual}, which is a non-linear rigid body model. A more detailed model is presented in \cite{cazzolato2011dynamics}. 

\textbf{Controller.} We consider a state feedback controller to stabilize the pendulum about the vertical equilibrium. The system has four states, $\x=[\alpha,\beta,\omega,\phi]$: the angular position of the rotary arm and the pendulum, $\alpha, \beta$, with $\beta=0$ being the vertical position, and the corresponding angular velocities, $\omega,\phi$. We control the voltage applied to the motor, $V_\mathrm{m}$. We use the encoder readings as  estimates of the angular positions, $\hat{\alpha}, \hat{\beta}$. We apply a low-pass filter to the difference of consecutive angular positions to estimate the angular velocities, $\hat{\omega}, \hat{\phi}$. We aim to find a controller of the form $u_t=V_\mathrm{m}=[\param_1, \param_2, \param_3, \param_4][\hat{\alpha}, \hat{\beta}, \hat{\omega}, \hat{\phi}]^\top$.

\textbf{Scaling and reward.} We define a state and action dependent reward as the negative of a quadratic cost, $r(\x, \ub)=-(\x^\top Q\x+\ub^\top R\ub)$ with $Q=\text{diag}(1, 10, 0, 0)$ and $R=8$. The performance associated to a controller is the expected average reward it induces on the system, that is, for a trajectory of duration $T$, $\mathbb{E}[1/T\int_0^T r(\x_t,\pi(\x_t| \theta))dt]$. To prevent one of the objectives from dominating the contribution to the hypervolume improvement in the EHI algorithm, we must normalize them. We control the range of the robustness indicators, see \cref{sec:robustness}, and, therefore it is easy to rescale them to the $[0, 1]$ range. We observe empirically that the unnormalized return ranges in $[-500, 0]$. Thus, we clip every return value to this range and we rescale it to the $[0, 1]$ interval. Since the pendulum incurs substantially different returns when a stabilizing or destabilizing controller is used, we cannot rescale the range linearly. Instead, we use a piece-wise linear function. In particular, since we observe empirically that stabilizing controllers have a performance between -20 and 0, we rescale linearly the range $[-500, -20]$ to $[0, 0.5]$ and the range $[-20, 0]$ to $[0.5, 1]$. This differentiates coarsely the quality of unstable controllers, and it gives a more refined scale over stable ones.

\begin{table*} 
	\caption{\simtoreal experiment: we train in simulation 5 different controllers with scalar BO, MOBO with delay margin or lower gain margin. We test each one 5 times on the hardware in 4 scenarios and we compare them according to 3 metrics (see the main text for an in depth description). The robust controllers consistently outperform the non-robust ones across all scenarios.}
	\label{tab:sim2real}
	\centering
	\setlength{\tabcolsep}{5pt}
	\footnotesize
	\begin{tabular}{l | c c c | c c c | c c c | c c c}
					 		& \multicolumn{3}{c}{Standard} & \multicolumn{3}{c}{Motor noise} & \multicolumn{3}{c}{Sensor noise}  & \multicolumn{3}{c}{Add $2\,\text{g}$} \\ 
					 		& $\mathbb{E}[R]$ & Fail & Fail time (s) & $\mathbb{E}[R]$ & Fail & Fail time (s) & $\mathbb{E}[R]$ & Fail & Fail time (s)& $\mathbb{E}[R]$ & Fail & Fail time (s) \\ 
		\toprule
		Scalar BO		& 		    -0.150		    & 	 80\%		  & 0.92			 &     -0.151				  &   80\% 		& 0.97 & 		    -0.151		    & 	 80\%		  & 1.05			 &     -0.185				  &   100\% 		& 1.03\\
		MOBO-DM	&			-0.044  		&  32\%   	  & 4.61 			 &     -0.038             &   20\% 		& 4.07  &			-0.063  		&  20\%   	  & 4.32 			 &    -0.126              &   84\% 		& 3.21\\
		MOBO-GM	&  \textbf{-0.003}  &\textbf{0\%}&$\boldsymbol{\infty}$& \textbf{-0.013}   &\textbf{0\%}& $\boldsymbol{\infty}$  &  \textbf{-0.057}  &\textbf{0\%}&$\boldsymbol{\infty}$    & \textbf{-0.004}   &\textbf{0\%}& $\boldsymbol{\infty}$\\
	\end{tabular}
\end{table*}
\begin{table*} [t]
	\caption{Hardware experiment: we train on hardware one controller with scalar BO and one with MOBO with lower gain margin. We test each one 5 times on the hardware in 4 scenarios and we compare them according to 3 metrics (see the main text for an in depth description). The robust controller consistently outperform the non-robust ones across all scenarios.}
	\label{tab:hardware}
	\centering
	\setlength{\tabcolsep}{5pt}
	\footnotesize
	\begin{tabular}{l | c c c | c c c | c c c | c c c}
					 		& \multicolumn{3}{c}{Add $5\,\text{g}$} & \multicolumn{3}{c}{Add $9\,\text{g}$} & \multicolumn{3}{c}{Add $10\,\text{g}$ and $8\,\text{cm}$} & \multicolumn{3}{c}{Add $5\,\text{g}$, motor + sensor noise} \\ 
					 		& $\mathbb{E}[R]$ & Fail & Fail time (s) & $\mathbb{E}[R]$ & Fail & Fail time (s) & $\mathbb{E}[R]$ & Fail & Fail time (s) & $\mathbb{E}[R]$ & Fail & Fail time (s) \\ 
		\toprule
		Scalar BO		& 	 		    -0.101		&\textbf{0\%}&$\boldsymbol{\infty}$&     -0.669				  &   100\% 		& 4.07   & 		    -0.711		    & 	 100\%		  &    3.59		  &   \textbf{-0.259} &\textbf{0\%}&$\boldsymbol{\infty}$\\
		MOBO-GM	&  \textbf{-0.031}  &\textbf{0\%}&$\boldsymbol{\infty}$& \textbf{-0.026}   &\textbf{0\%}& $\boldsymbol{\infty}$    &   \textbf{-0.0259}  &\textbf{0\%}&$\boldsymbol{\infty}$& -0.366       &\textbf{0\%}& $\boldsymbol{\infty}$\\
	\end{tabular}
	\vspace{-.3cm}
\end{table*}

\textbf{Surrogate models.} For the non-robust algorithm, we use a standard GP model with a zero mean prior and a Mat\'{e}rn kernel with $\nu=5/2$ with automatic relevance determination (ARD). We set the hyperprior over the lengthscales to $\mathrm{Lognormal}(1, 3)$ and over the standard deviation to $\mathrm{Lognormal}(0.35, 1)$. We use a Gaussian likelihood with no hyperprior. Similarly, for the robust algorithm, we use a zero prior mean. The correlation in the input space in the ICM model is captured by an ARD Mat\'{e}rn kernel with $\nu=5/2$, with the same hyperprior as the non-robust case. For the correlation in the output space, we set a Gaussian hyperprior over each entry of the matrix $\mathbf{B}$, $\mathcal{N}(0, 1)$. We use a Gaussian likelihood with a diagonal covariance matrix. In both cases, we udpate the hyperparameters using a maximum a posteriori estimate after every new data point is acquired.



\textbf{Training}. In the \simtoreal setting, we train 5 different controllers for each of these methods: scalar BO (non-robust), MOBO with performance and delay margin (DM), and MOBO with performance and lower gain margin (GM). The training consists of 200 BO iterations evaluated in simulation. In the hardware training setting, we train one controller for scalar BO and one for MOBO-GM using 70 BO iterations evaluated on hardware. In both settings, MOBO requires fewer iterations than the given budget to find satisfactory solutions. Thus, using a stopping criterion in \cref{alg:EHI} would reduce the total number of iterations.
We estimate performance by averaging the return over 10 independent runs. To estimate robustness, we require that the controller stabilizes the system for a given delay or gain for 5 independent runs. A trial is deemed stable if $\alpha_t \in [-8^\circ,8^\circ]$  and $\beta_t \in [-4^\circ,4^\circ]$ for all $t\in [4,5]$. Every training run lasts for 5 seconds. \cref{fig:dm_front,fig:gm_front} show the fronts obtained by the MOBO-DM and MOBO-GM  \simtoreal training, respectively. The gray circles correspond to controllers that appeared stabilizing at first, but that were ruled out with longer simulations, cf.\ \cref{sec:robustness}. The green squares indicate controllers tested on hardware. To emphasize the generality of our method, they were selected to be approximately at the elbow of the front without further tuning.

\textbf{Sim-to-real test.} We test each controller learned in simulation on the hardware 5 times in 4 scenarios: (\emph{i}) standard sim-to-real, (\emph{ii}) \simtoreal adding Gaussian noise $\mathcal{N}(0, 0.5)$ to the motor voltage, (\emph{iii}) \simtoreal  adding noise to the encoder readings following a multinomial distribution over the integers in $[-4, 4]$ with $p(0)=0.6$ and 0.05 everywhere else, and (\emph{iv}) \simtoreal with the pendulum mass increased by $\SI{2}{\gram}$. A run is a failure if $|\beta|>20^\circ$. In \cref{tab:sim2real}, we compare the controllers in terms of average return, failure rate, and failing time, averaged over the runs that resulted in a failure. The robust methods consistently outperform the non-robust policy optimization across all test scenarios. It appears that the lower gain margin is a more suitable robustness indicator in this setting. This may be due to the fact that, in our experience, the gain margin is less noisy to estimate.

\textbf{Hardware test.} We test each controller learned in hardware 5 times in 4 scenarios: (\emph{i}) extra mass of \SI{5}{\gram}, (\emph{ii}) extra mass of \SI{9}{\gram}, (\emph{iii}) extra mass of \SI{10}{\gram} and extra pendulum length of \SI{8}{\cm}, and (\emph{iv}) extra mass of \SI{5}{\gram} with the actuation and sensor noise used in the \simtoreal experiments. \cref{tab:hardware} summarizes the test results. Similarly to the \simtoreal setting, the robust algorithm consistenlty outperforms its non-robust counterpart.


%% file: sections/6-conclusions.tex
\section{Concluding Remarks}
\label{sec:conclusion}
We present a data-efficient algorithm for robust policy optimization based on multi-objective Bayesian optimization. We suggest a data-driven evaluation of two common robustness indicators, which is suitable to model-free settings. Our hardware experiments on a Furuta pendulum show that \emph{(i)} our method facilitates simulation to real transfer, and \emph{(ii)} consistently increases robustness of the learned controllers 
as compared to BO with a single performance objective.
%
Our results indicate a promising avenue toward robust learning control by leveraging robustness measures from control theory and multi-objective Bayesian optimization and
 point to several directions for extensions.  
 While we show that gain and delay margings are effective in practice on a mildly nonlinear system, they may not fully characterize robust stability in general \cite{zhou1996robust,boulet2007fundamental}.
Thus, investigating other relevant robustness indicators that can efficiently be estimated from data in a model-free setting is a topic for future research. Also, using multiple robustness indicator simultaneously is relevant, which our method could do at the expense of a more complex scaling to balance robustness and performance.

\clearpage